\documentclass[sigconf]{acmart}

\usepackage{graphicx} 
\usepackage[english]{babel}
\usepackage{blindtext}
\usepackage{todonotes}
\usepackage[absolute,overlay]{textpos}

\AtBeginDocument{%
  \providecommand\BibTeX{{%
    \normalfont B\kern-0.5em{\scshape i\kern-0.25em b}\kern-0.8em\TeX}}}

\copyrightyear{2019} 
\acmYear{2019} 
\setcopyright{acmlicensed}
\acmConference[Embedded Systems Week 2019]{INTelligent Embedded Systems Architectures and Applications Workshop 2019}{October 13--18, 2019}{New York, NY, USA}
\acmBooktitle{INTelligent Embedded Systems Architectures and Applications Workshop 2019 (INTESA2019), October 13--18, 2019, New York, NY, USA}
\acmPrice{15.00}
\acmDOI{10.1145/3372394.3372396}
\acmISBN{978-1-4503-7652-5/19/10}

\begin{document}

\title{An End-to-End HW/SW Co-Design Methodology to Design Efficient Deep Neural Network Systems using Virtual Models}
\renewcommand{\shorttitle}{End-to-End HW/SW Co-Design Methodology to Design Efficient DNN Systems using Virtual Models}

\author{Michael J. Klaiber}
\email{michael.klaiber@de.bosch.com}
\orcid{1234-5678-9012}
\affiliation{
  \institution{Robert Bosch Corporate Research}
  \city{Renningen}
  \country{Germany}
  \postcode{71272}
}

\author{Sebastian Vogel}
\email{sebastian.vogel@de.bosch.com}
\affiliation{
  \institution{Robert Bosch Corporate Research}
  \city{Renningen}
  \country{Germany}
  \postcode{71272}
}

\author{Axel Acosta}
\email{axel.acosta@de.bosch.com}

\affiliation{
  \institution{Robert Bosch Corporate Research}
  \city{Renningen}
  \country{Germany}
  \postcode{71272}
}

\author{Robert Korn}
\email{robert.korn@de.bosch.com}

\affiliation{
  \institution{Robert Bosch Corporate Research}
  \city{Renningen}
  \country{Germany}
  \postcode{71272}
}

\author{Leonardo Ecco}
\email{leonardo.ecco@de.bosch.com}

\affiliation{
  \institution{Robert Bosch Corporate Research}
  \city{Renningen}
  \country{Germany}
  \postcode{71272}
}

\author{Kristine Back}
\email{kristine.back@de.bosch.com}

\affiliation{
  \institution{Robert Bosch Corporate Research}
  \city{Renningen}
  \country{Germany}
  \postcode{71272}
}

\author{Andre Guntoro}
\email{andre.guntoro@de.bosch.com}

\affiliation{
  \institution{Robert Bosch Corporate Research}
  \city{Renningen}
  \country{Germany}
  \postcode{71272}
}

\author{Ingo Feldner}
\email{ingo.feldner@de.bosch.com}

\affiliation{
  \institution{Robert Bosch Corporate Research}
  \city{Renningen}
  \country{Germany}
  \postcode{71272}
}

\renewcommand{\shortauthors}{M. J. Klaiber et al.}

\begin{abstract}
  End-to-end performance estimation and measurement of deep neural network (DNN) systems become
  more important with increasing complexity of DNN systems consisting of hardware and software components.
  The methodology proposed in this paper aims at a
  reduced turn-around time for evaluating different design choices of hardware and software components of DNN systems.
  This reduction is achieved by moving the performance estimation from
  the implementation phase to the concept phase by employing virtual hardware models
  instead of gathering measurement results from physical prototypes.
  Deep learning compilers introduce hardware-specific transformations and are, therefore, considered a part of the design flow
  of virtual system models to extract end-to-end performance estimations.  
  To validate the run-time accuracy of the proposed methodology, a system processing the \emph{DilatedVGG}
  DNN is realized both as virtual system model and as
  hardware implementation. The results show that up to 92 \% accuracy can be reached
  in predicting the processing time of the DNN inference.

\end{abstract}




\maketitle

\begin{textblock*}{15cm}(3cm,1cm)
  \begin{center}
  \begin{tiny}
  \textcopyright ACM 2019 This is the author's version of the work. It is posted here for your
  personal use. Not for redistribution. The definitive version was published in \\
  Embedded Systems Week 2019,
  INTelligent Embedded Systems Architectures and Applications Workshop 2019, https://doi.org/10.1145/3372394.3372396
  \end{tiny}
  \end{center}
\end{textblock*}

\section{Introduction and Motivation}
Systems to perform the inference phase of a learned deep neural networks (DNN) have become more important in many fields from automotive to Industry 4.0 applications.
In the beginning of neural network research
there were a limited number of computational kernels which
lend themselves to simple hardware structures for efficient execution.
With the emerge of graphics processing units (GPUs) deeper neural network structures were
feasible to train and deploy, as they were able to efficiently map those computational kernels.
Modern DNNs, however, have become more complex in structure and operations, e.g. InceptionNet-v4 or NasNET \cite{Szegedy2017, Zoph2018}.
Future DNNs are expected to become even more complex in structure and the number of different operations, e.g. DNNs generated by the uprising field of \emph{neural architecture search} \cite{Zoph2018, Elsken2018}.
Designing systems to process the current and future DNNs requires, therefore, an efficient and powerful methodology to
balance compute and communication resources.

\begin{figure}[t]
    \includegraphics[width=\columnwidth]{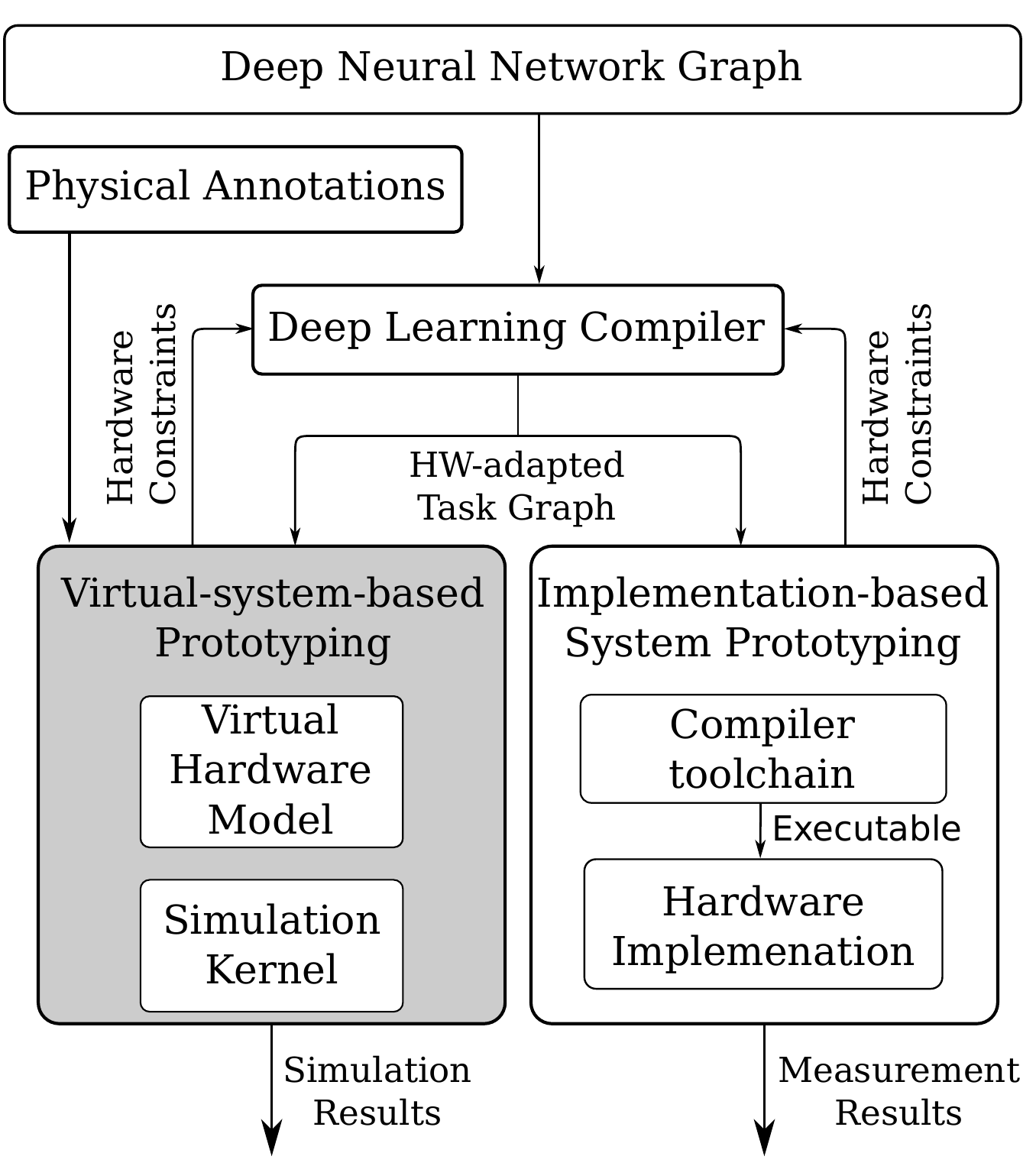} 
  \caption{Similarities and differences of implementation-based and virtual-system-based prototyping flows. \label{fig:flow_and_architecture_a}}
\end{figure}

Models of hardware or software components that only mimic the timing behavior and the memory transactions of a component
while neglecting functional computation are referred to as non-functional virtual models.
The used methodology is in essence similar to transaction-level modeling.
All models introduced in the
following are implicitly non-functional.
Virtual hardware models
are executable high-level descriptions of hardware elements (e.g. CPUs, interconnects or memories) that annotate operations with
simulation cycles.
A deep learning compiler breaks the DNN graph down into a graph where each node represents a memory access or
processing cycles on a virtual hardware model. This graph is called the \emph{Task Graph},
and is effectively a virtual software model.
A combination of multiple virtual
hardware models (e.g. shown in Figure \ref{fig:flow_and_architecture_a}) and a task graph is referred to as an abstract virtual system model (AVSM).
Due to the high abstraction level of an AVSM, these models are much faster compared to  a simulation at
register-transfer level (RTL) and allow to model system aspects,
e.g. timing behavior within a tolerable error.

In a DNN system, the requirements on processing elements, as well as the requirements on
communication infrastructure, are mainly driven by
the topology and the arithmetic operations of the target DNNs.
When trying to evaluate novel concepts for DNN systems
that consist both of hardware and software components, an accepted approach is to implement a prototype
and measure the performance of the target DNNs.
This requires development of software and hardware components (and possibly manufacturing of hardware) for the evaluation of one specific concept.
The huge design space for DNN systems in the algorithmical domain, the software domain and the hardware domain,
makes finding and evaluating efficient concepts considerably time consuming.

There are also analytical approaches \cite{Zhang2015, Motamedi2016, Yu2019} for designing DNN systems, e.g. by analyzing bandwidth and computation requirements of the DNN  and applying techniques like loop tiling or transformation
\cite{Zhang2015}.
However, deep learning compilers that transform the DNN graph, hence create the task graph, are
often neglected in this optimization process.
Some approaches use statistical methods for performance estimation.
They use the frequency at which events of previously known communication patterns occur to describe a system,
whereas simulation considers the causality. Therefore, simulation is more adequate to detect communication bottlenecks and blocking behavior.

The lack of an end-to-end methodology that considers both hardware architecture and software tool chain
becomes particularly apparent by the enormous
number of 
publications that describe 
implementations of DNN systems for one specific design point. 

In this paper, we address these challenges by proposing a sim\-u\-la\-tion-based end-to-end methodology that uses virtual hardware models in combination with a deep learning compiler to evaluate the performance of novel concepts for DNN systems.

\begin{figure}[t]
    \includegraphics[width=\columnwidth]{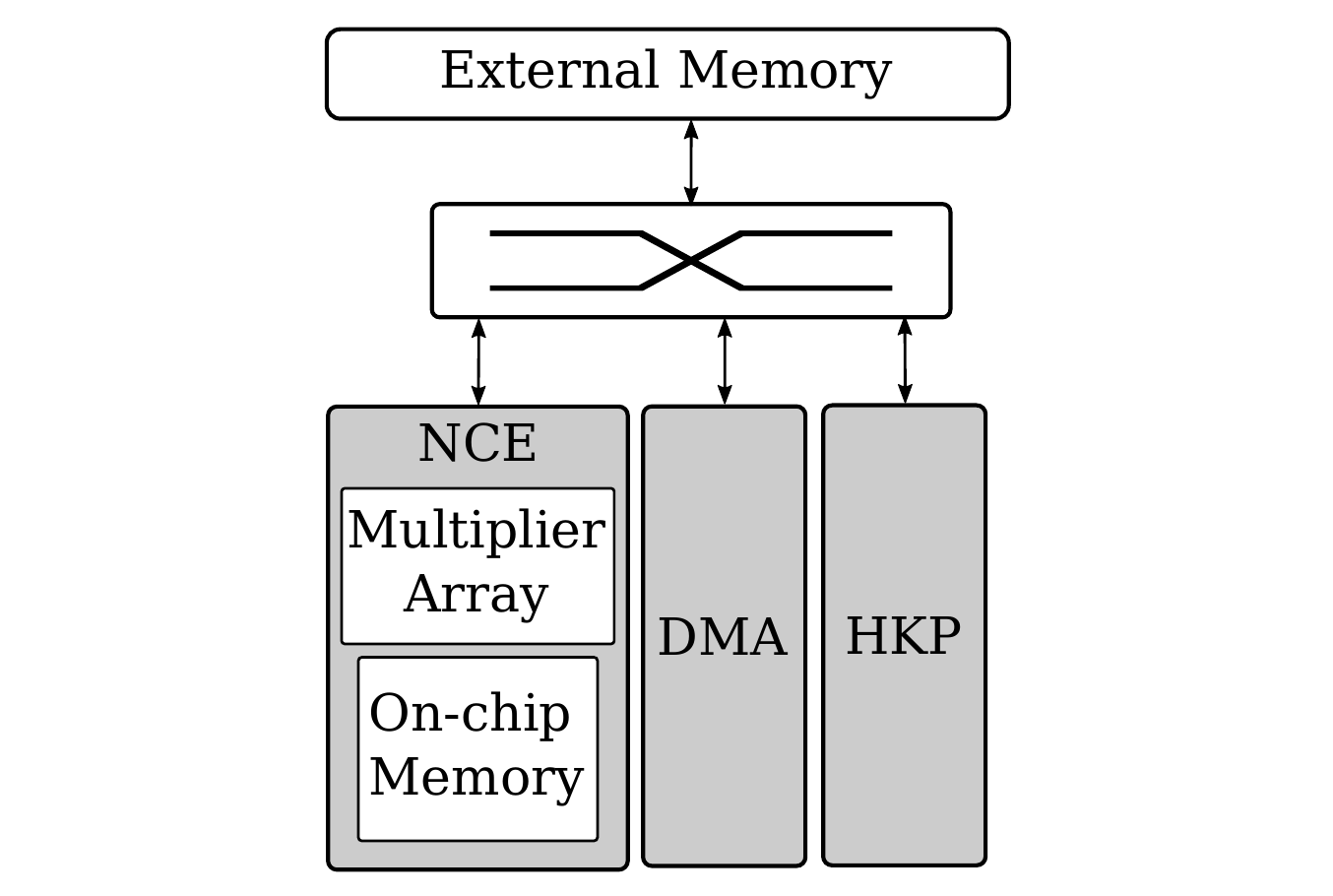}
  \caption{Base architecture of DNN system. \label{fig:flow_and_architecture_b}}
\end{figure}

\section{Methodology}

Evaluating novel concepts for DNN systems by implementing physical prototypes requires
a full iteration in a hardware and software development cycle for only a single design point.
For totally new hardware concepts, designing and fully implementing an extension of the tool chain
might be even necessary.
This significantly limits the number of iterations in the development cycle within the scope of a project, i.e. limits the design space that can be explored.
Figure \ref{fig:flow_and_architecture_a} shows the implementation-based prototyping flow,
as well as the virtual-system-based prototyping flow.

The DNN system architecture shown in Figure \ref{fig:flow_and_architecture_b} is based on the common properties of state-of-the-art DNN systems 
and, therefore, serves as starting point to evaluate the methodology proposed in the following.
Feature maps, weights and intermediate data are stored in external memory. The matrix multiplications
and other arithmetic operations are performed by the \emph{Neural Complex Engine} (NCE). Memory transactions are
carried out by a direct memory access (DMA) hardware. All components of the system are connected by an interconnect
and controlled by a house-keeping processor (HKP) that executes the task graph.
The flows from Figure \ref{fig:flow_and_architecture_a}(a) are exemplified for the DNN system architecture shown in Figure \ref{fig:flow_and_architecture_b}.

In both flows, the deep learning compiler converts the DNN graph to a hardware-adapted
task graph according to the
hardware constraints that are either provided by the virtual hardware model or the hardware implementation.
The resulting task graph considers the memory hierarchy, the on-chip memory sizes
and the supported operations of the DNN system.

The implementation-based prototyping flow requires an implementation of all the hardware at RTL level. To measure/evaluate the performance, there are two possibilities,
either a simulation on RTL level or a performance analysis after manufacturing. The RTL level simulation has the disadvantage that running a single inference of a
DNN requires several hours or days depending on the DNN's complexity. Manufacturing the system, of course,
provides the most accurate measurement results, but has a significantly slower turn-around time.

By contrast our virtual-system-based prototyping flow requires virtual hardware models for all hardware components.
Compared to an RTL implementation, these components are described at a higher level which results in faster
implementation time. To determine the system performance, the task graph is deployed in the virtual model of
the HKP which controls the execution of the virtual

model of the NCE.

In the implementation-based flow careful engineering is required to meet physical constraints.
A virtual-system-based approach aims at faster evaluation by modeling a high-level system description.
Therefore, physical annotations, such as clock frequency, are imported to the AVSM.

The described properties enable the AVSM methodology to assess the performance either in bottom-up or in a
top-down manner.
If the DNN system's target performance is known, it is possible to assess physical requirements
(e.g. the required frequency) of components, such as for the NCE.
For the case where physical annotation of a component are already available,
the performance and scalability at system level can be estimated accurately.

\section{Preliminary Results}

The flow outlined in Figure \ref{fig:flow_and_architecture_a} is implemented as Python framework we developed specially for the purpose of modeling abstract virtual system models (AVSMs).
This framework consists of a library of parametrizable components to describe hardware components,
a compiler interface to transform the internal graph representation into a hardware-adapted task graph,
and a model generation engine.

Each instance of an AVSM is described as \emph{system description file} that defines the topology of
the virtual hardware models of the NCE, the memory sub-system and the bus.
It also contains the physical annotations, such as the frequency of the NCE or the memory frequency.
The model generation engine then uses the system description file and the hardware-adapted task graph
to automatically generate an executable \emph{SystemC} model that is 
simulated in Synopsys Platform Architect.

\begin{figure}[t]

    \includegraphics[width=\columnwidth]{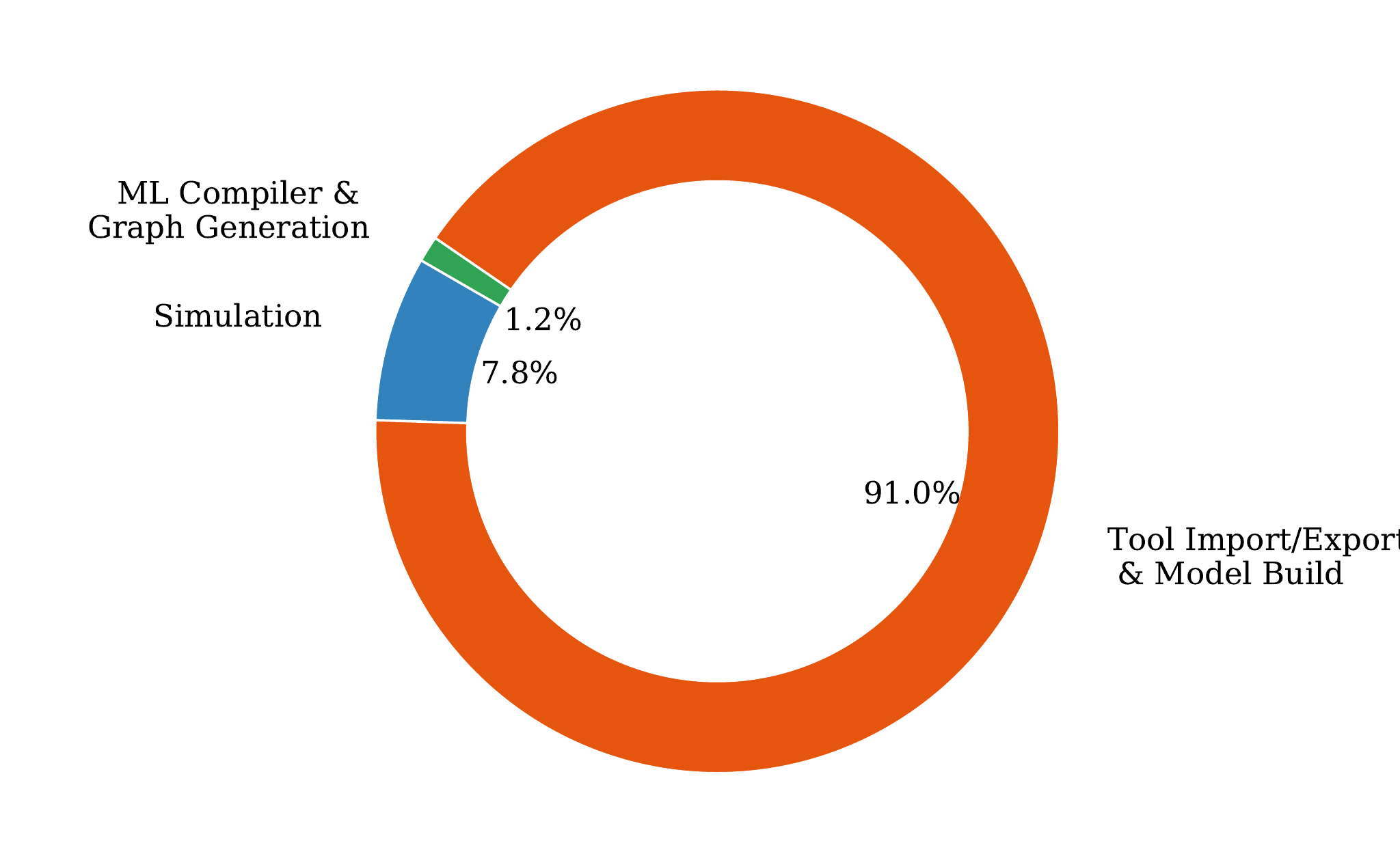}
    \begin{tabular}{|c|c|}
      \hline
      Task & Time [s] \\
      \hline \hline
      Simulation & 105.82 \\
      \hline
      Tool import/export and Model build & 1231.08\\
      \hline
      ML Compiler \& Graph Generation &	16.64 \\
      \hline
      $\sum$ & 1353.54\\
      \hline
    \end{tabular}
    
    \caption{Distribution of run-time for generation and simulation of ASVM. \label{fig:run_time_chart}}
\end{figure}

\begin{figure}
  \includegraphics[width=\columnwidth]{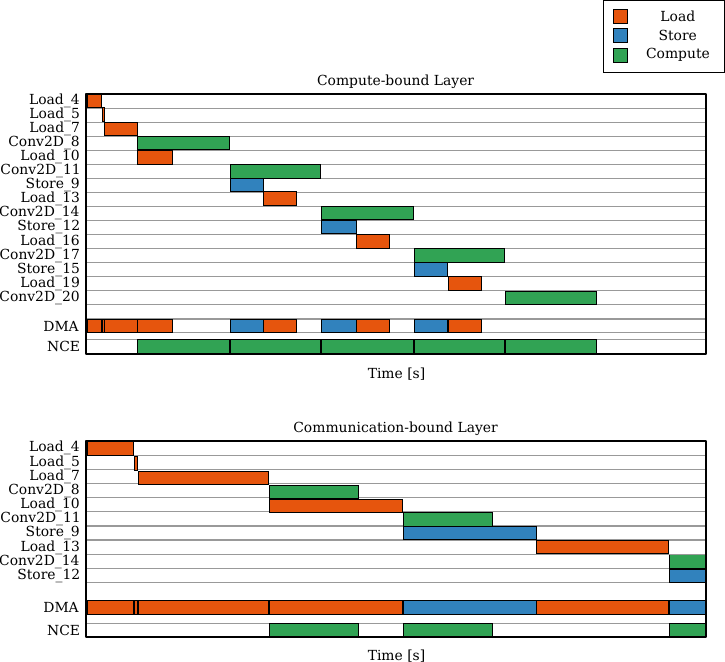}
    \caption{Gantt chart showing simulation of tasks and usage of computation and communication resources. \label{fig:resource_chart}}
\end{figure}

As run-time is one of the major advantages of the presented virtual-system-based prototyping flow, 
Figure \ref{fig:run_time_chart} shows the total run-time to build an AVSM from the
\emph{system description file} and simulate all layers of the \emph{DilatedVGG}
neural network.
The total processing time on an Intel Xeon CPU E5620 running at 2.40 GHz is around 20 minutes.
Generation of the hardware-adapted task graph and the hardware models of the AVSM takes 16.4 seconds.
The simulation of the resulting SystemC model is carried out in 105 seconds.
A majority of approximately 91\% of the processing time is spent for importing the hardware-adapted task graph,
exporting the results and the build process of the SystemC model. This part of the flow has not been optimized for
performance yet, therefore, bears a great potential for further improvement.

\begin{figure*}
  \includegraphics[width=0.9\textwidth]{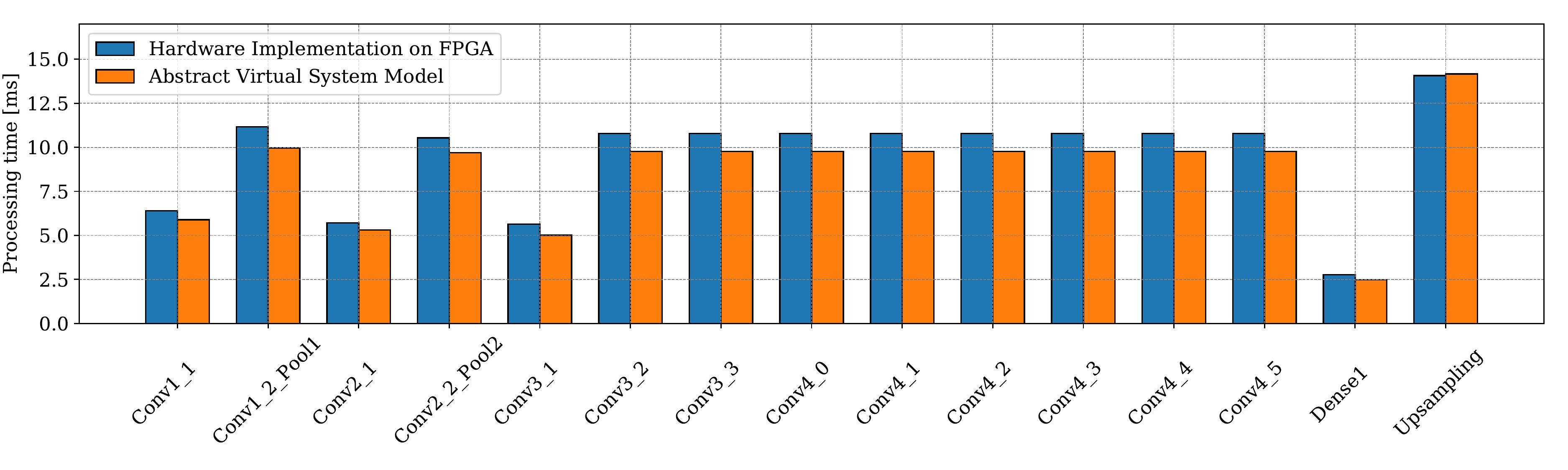}
  
    \caption{Comparison of HW implementation and AVSM. \label{fig:comparison_graph}}
\end{figure*}

To compare the presented methodology quantitatively, an AVSM and an FPGA implementation
of the DNN system architecture shown in Figure \ref{fig:flow_and_architecture_b} were created.
The physical prototype \cite{Vogel2019} was implemented on a Xilinx Virtex7 FPGA platform, with an NCE
that has $32 \times 64$ multipliers 
that run at a clock frequency of
\emph{250 MHz}.
Figure \ref{fig:comparison_graph} compares the processing time of the physical prototype and
the AVSM for processing a (slightly modified version)
of the \emph{DilatedVGG} DNN \cite{Yu2015}.
The total processing time for a single inference deviates by 8.3\%
when comparing the physical prototype and the AVSM.
Individual layers deviate between 0.6 \% and 11.2 \%. This deviation is a result of a high-level
model of the memory sub-system and could be further improved by adding more
hardware properties into the virtual hardware models.

The virtual-system-based prototyping allows to track computation time at the level of individual operations and
the traffic on the bus for each memory transaction.
Therefore, a detailed analysis of the performance and efficiency
for a design point of a DNN system is possible.
The Gantt chart in Figure \ref{fig:resource_chart} shows the dependencies of
memory transactions and the computations, as well as the usage of the
communication resources and the computation resources for communication-bound and compute-bound layers.
For the compute-bound bound layers, the hardware model of the NCE is occupied continuously, the
hardware model of the DMA is partially vacant; in the compute-bound case this is the other way around.
The detailed level of observability of the AVSM, therefore, allows a virtual performance analysis for
each layer to identify potential performance bottlenecks.

\begin{figure}[t]
  \includegraphics[width=\columnwidth]{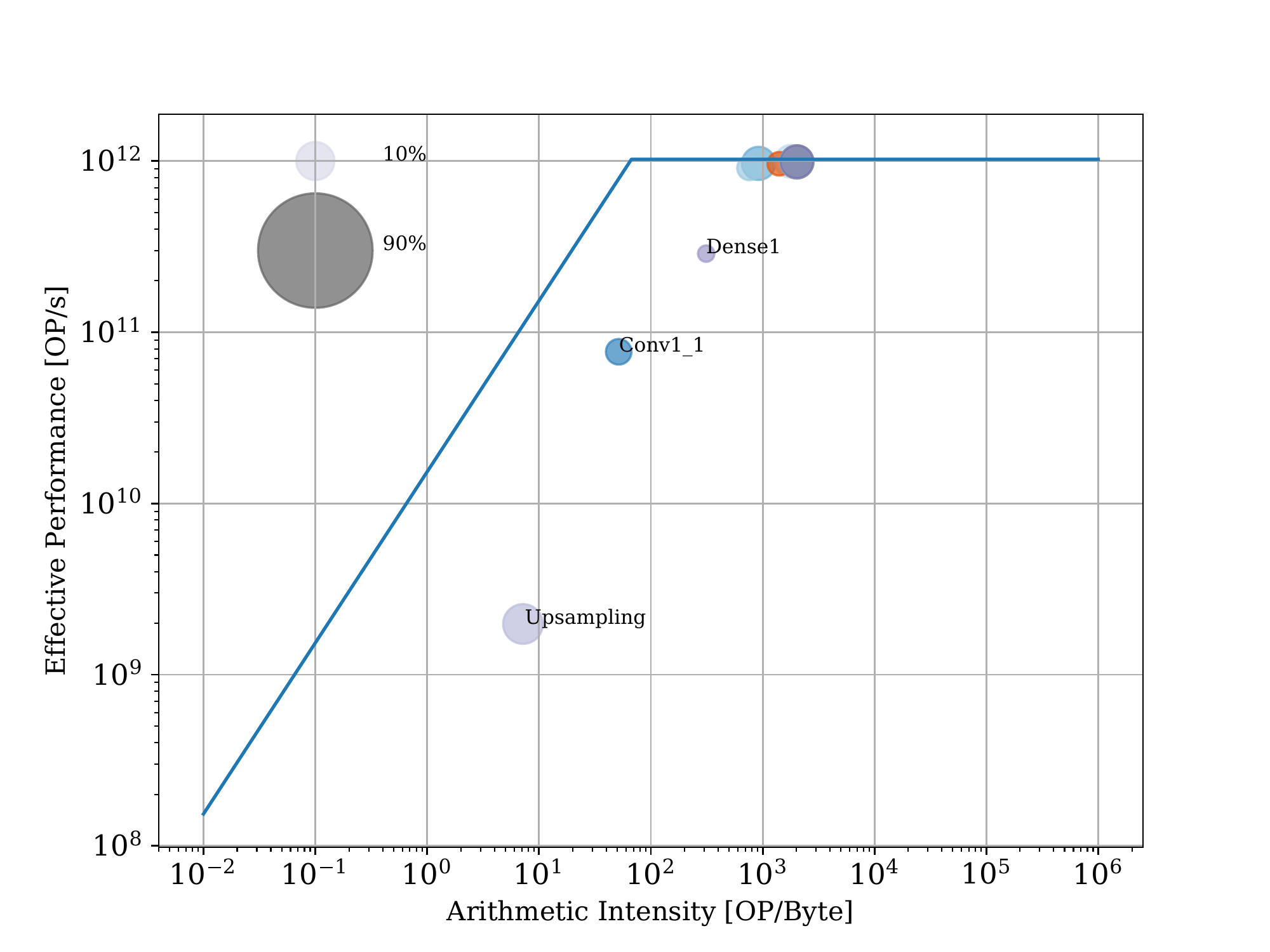}
    \caption{Roofline model of AVSM executing \emph{DilatedVGG}. \label{fig:roofline_chart}}
\end{figure}

\begin{figure}
  \includegraphics[width=\columnwidth]{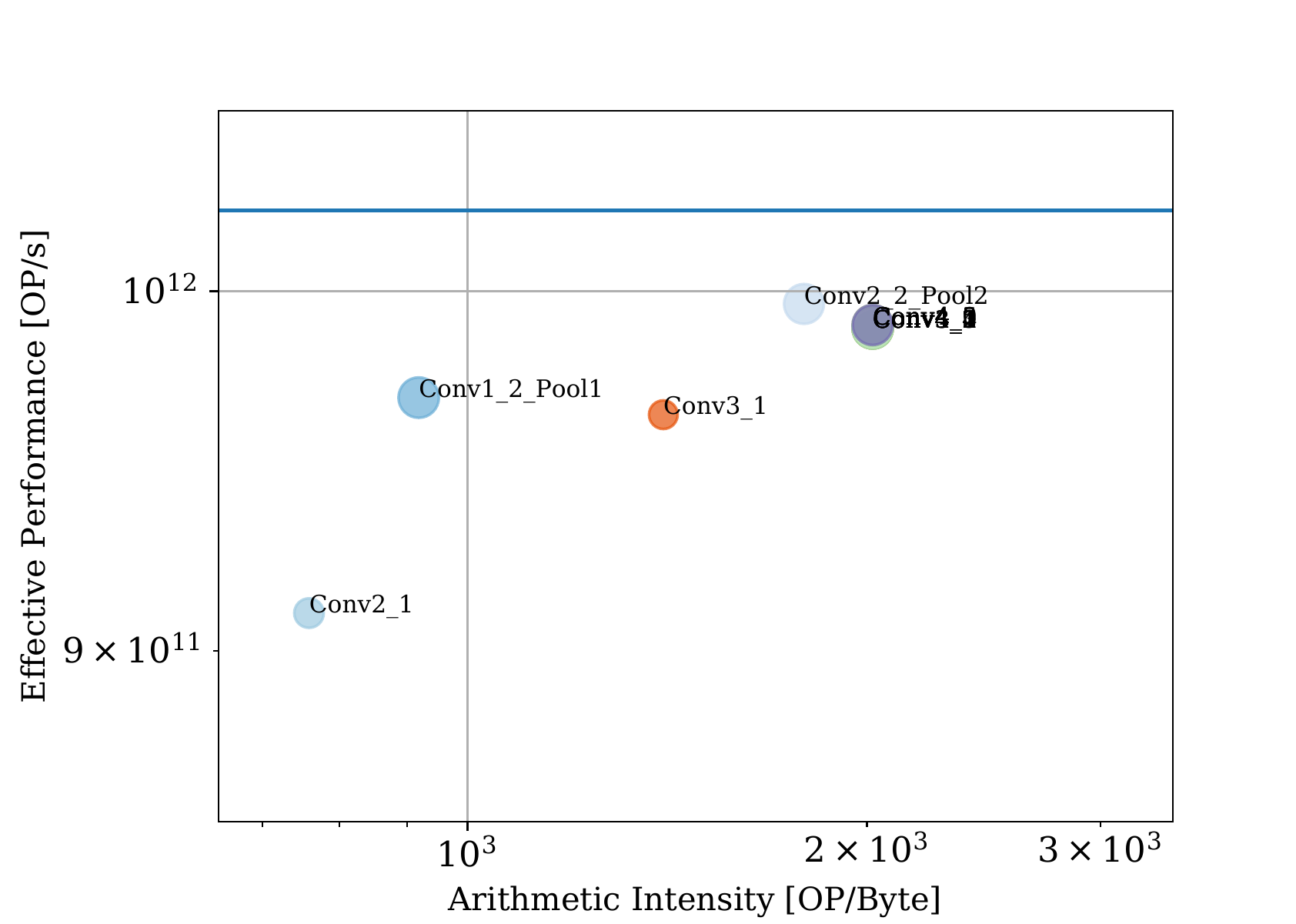}
    \caption{View on compute-bound layers in Figure \ref{fig:roofline_chart}. \label{fig:roofline_chart_zoomed}}
\end{figure}

The roofline model \cite{Williams2009} in Figure \ref{fig:roofline_chart} visualizes the performance and efficiency of the
AVSM of the DNN system specified in the previous paragraph for processing each layer of the \emph{DilatedVGG} DNN.
The layers are represented by dots,
where the size of a dot shows the execution time in relation to the time required for a single inference of
the neural network. Most layers are fairly close to the vertical threshold of the roofline (e.g. \emph{Conv4\_0} -- \emph{Conv4\_5}),
which indicates that these layers are compute-bound. Figure \ref{fig:roofline_chart_zoomed}
zooms into the part of Figure \ref{fig:roofline_chart} to show the compute-bound layers.
This indicates that increasing the peak performance of the DNN could accelerate the processing of these layers.
The layers \emph{Dense1}, \emph{Upscaling}, \emph{Conv1\_1} are neither compute-bound, nor communication-bound.
Therefore, increasing the peak performance or the bus bandwidth of the DNN system would not necessarily have an
effect on their execution time. Possibilities for accelerating those layers range from 
software approaches like changing how the Deep Learning compiler maps and transforms individual operations, 
to optimizations of low level hardware like the arrangement of the multiplier array or the hierarchy of the on-chip memory in the NCE.
Both, the software and the hardware changes can be done in the AVSM. This underlines the importance of an end-to-end flow
for optimizing DNN systems.

\clearpage

\section{Conclusion}
The presented results show that AVSMs are a promising alternative
to classical implementation-based physical prototypes.
Compared to an implementation at RTL level, the turn-around time for generating performance
estimations of DNN systems is significantly faster with AVSMs.
The end-to-end design space exploration of DNN systems in a huge design space can easily be done by a \emph{click of a button}.
The tight coupling of the deep learning compiler
to the AVSMs in the proposed methodology provides accurate results with less than 9 \%  deviation
for the evaluated cases.

\bibliographystyle{ACM-Reference-Format}
\bibliography{intesa}

\end{document}